\documentclass[conference]{IEEEtran}
\usepackage{amsfonts}
\usepackage{times} 

\usepackage{amssymb,amsmath,amsthm}
\usepackage{graphicx,epstopdf,color}
\usepackage{caption}
\usepackage{subcaption}
\usepackage{wrapfig}
\usepackage{algorithm}
\usepackage{algorithmicx}
\usepackage{algpseudocode}
\algdef{SE}[DOWHILE]{Do}{doWhile}{\algorithmicdo}[1]{\algorithmicwhile\ #1}

\ifCLASSINFOpdf
\else
\fi

\IEEEoverridecommandlockouts    

\begin{document}
%

\title{\ \\ \LARGE\bf Sub-Classifier Construction for Error Correcting Output Code Using Minimum Weight Perfect Matching\thanks{Patoomsiri Songsiri and Boonserm Kijsirikul are with the Department of Computer Engineering, Chulalongkorn University, Bangkok, Thailand (email: patoomsiri.s@student.chula.ac.th and boonserm.k@chula.ac.th).}
\thanks{Thimaporn Phetkaew is with the School of Informatics, Walailak University, Nakhon Si Thammarat, Thailand (email: pthimapo@wu.ac.th).}
\thanks{Ryutaro Ichise is with the National Institute of Informatics, Tokyo, Japan (email: ichise@nii.ac.jp).}
}

\author{
{Patoomsiri Songsiri, Thimaporn Phetkaew, Ryutaro Ichise and Boonserm Kijsirikul }
}

\maketitle

\begin{abstract}
Multi-class classification is mandatory for real world problems 
and one of promising techniques for multi-class classification is Error Correcting Output Code. 
We propose a method for constructing the Error Correcting Output Code to obtain 
the suitable combination of positive and negative classes encoded to represent binary classifiers.
The minimum weight perfect matching algorithm is applied to find the optimal pairs of subset of classes
by using the generalization performance as a weighting criterion.
Based on our method, each subset of classes with positive and negative labels is appropriately combined for learning the binary classifiers.
Experimental results show that our technique gives significantly higher performance compared to traditional methods
including the dense random code and the sparse random code both in terms of accuracy and classification times.
Moreover, our method requires significantly smaller number of binary classifiers while maintaining accuracy compared to the One-Versus-One.

\end{abstract}

\begin{IEEEkeywords}
multi-class classification ; error correcting output code ; minimum weight perfect matching ; generalization performance
\end{IEEEkeywords}

\section{Introduction}

Error Correcting Output Code (ECOC)~\cite{Dietterich95,Allwein2000} is one of the well-known techniques for solving multiclass classification.
Based on this framework, an unknown-class instance will be classified by all binary classifiers 
corresponding to designed columns of the code matrix, and then the class with the closet codeword is assigned to the final output class. 
Each individual binary function including the large number of classes indicates the high capability as a shared classifier. 
Generally, when the number of classes increases, 
the complexity for creating the hyperplane also increases. 
The suitable combination of classes with the proper number of classes 
for constructing the model is still a challenge issue to obtain the effective classifier.    

Several classic works are applied to the design of the code matrix such as 
One-Versus-One (OVO)~\cite{Hastie98}, One-Versus-All (OVA)~\cite{Vapnik98}, 
dense random code, and sparse random code~\cite{Dietterich95,Allwein2000},
and for an $N$ class problem, they provide the number of binary models of $N(N-1)/2$, $N$, $\lceil 10log_2N \rceil$,  
and $\lceil 15log_2N \rceil$, respectively. Moreover, some approaches using the genetic algorithm have been 
proposed~\cite{Kuncheva2005,Lorena2009}. However, according to the complexity of the problem 
that the solutions are searched from a large space including all possible $2^{N-1}-1$ 
columns~\cite{Dietterich95} in case of dense code, and $\frac{3^{N}-2^{N+1}+1}{2}$ columns~\cite{Bagheri2013} 
in case of sparse code, design of code matrix is still ongoing research.

This research aims to find the suitable combination of classes for creating the binary models 
in the ECOC framework by providing both good classification accuracy and the small number of classifiers.
Our method is based on the minimum weight perfect matching algorithm 
by using the relation between pair of subset of classes defined by the generalization performance 
as the criterion for constructing the code matrix.
We study this multiclass classification based on Support Vector Machines~\cite{Vapnik95,Vapnik99} as base learners.
We also empirically evaluate our technique by comparing with the traditional methods 
{on ten datasets from the UCI Machine Learning Repository~\cite{Blake98}.
}

This paper is organized as follows. Section~\ref{ECOC_framework} reviews the traditional ECOC frameworks.
Section~\ref{Proposed_Method} presents our proposed method.
Section~\ref{experiments} performs experiments and explains the results and discussions.
Section~\ref{conclusion} concludes the research and directions for the future work.

\section{Error Correcting Output Codes}
\label{ECOC_framework}

Error Correcting Output Code (ECOC) was introduced by Dietterich and Bakiri~\cite{Dietterich95}
as a combining framework for multiclass classification. 
For a given code matrix with $N$ rows and $L$ columns, each element contains either 
$\textquoteleft$1', or $\textquoteleft$-1'.
This code matrix is designed to represent a set of $L$ binary learning problems
for $N$ classes. Each specific class is defined by the unique bit string called {\it{codeword}}
and each sub-problem is indicated by the combination of positive and negative classes 
corresponding to the elements of the code matrix.
Moreover, in order to allow the binary model learned without considering some particular classes,
Allwein et al.~\cite{Allwein2000} extended the coding method by adding the third symbol $\textquoteleft$0'
as {\it`` don't care bit''}. Unlike the previous method, the number of classes for training a binary classifier 
can be varied from 2 to $N$ classes. 

Several classic coding designs have been proposed, e.g., 
One-Versus-All (OVA)~\cite{Vapnik98}, dense random codes, sparse random codes~\cite{Allwein2000},
and One-Versus-One (OVO)~\cite{Hastie98}; the first two techniques and the last two techniques
are binary and ternary strategies, respectively. 
One-Versus-All codes including $N$ columns were designed by setting the $i^{th}$ class, 
and the remaining classes labeled with the positive and negative classes, respectively.
Dense random codes and sparse random were introduced by randomizing sets of 
$\lceil 10log_2N \rceil$ and $\lceil 15log_2N \rceil$ binary functions, respectively.
In case of One-Versus-One codes, they were designed to define each column 
by labeling $\textquoteleft$1' and $\textquoteleft$-1' to only two out of $N$ classes, 
and therefore there are $N(N-1)/2$ possible columns.

For a decoding process, an instance with unknown-class will be classified by all $L$ binary functions 
corresponding to designed columns of the code matrix. This output vector is compared to 
each row of the code matrix. The class corresponding the row of code matrix 
that provides highest similarity is assigned as the final output class.
Several similarity measures have been proposed such as Hamming distance~\cite{Dietterich95}, 
Euclidean distance~\cite{Escalera2010}, extended strategies based on these two methods
~\cite{Escalera2010}, and the loss-based technique~\cite{Allwein2000}.

\section{Proposed Method}
\label{Proposed_Method}

\begin{figure}[htbp]
\hspace{-35pt}
\scalebox{0.33}{\includegraphics{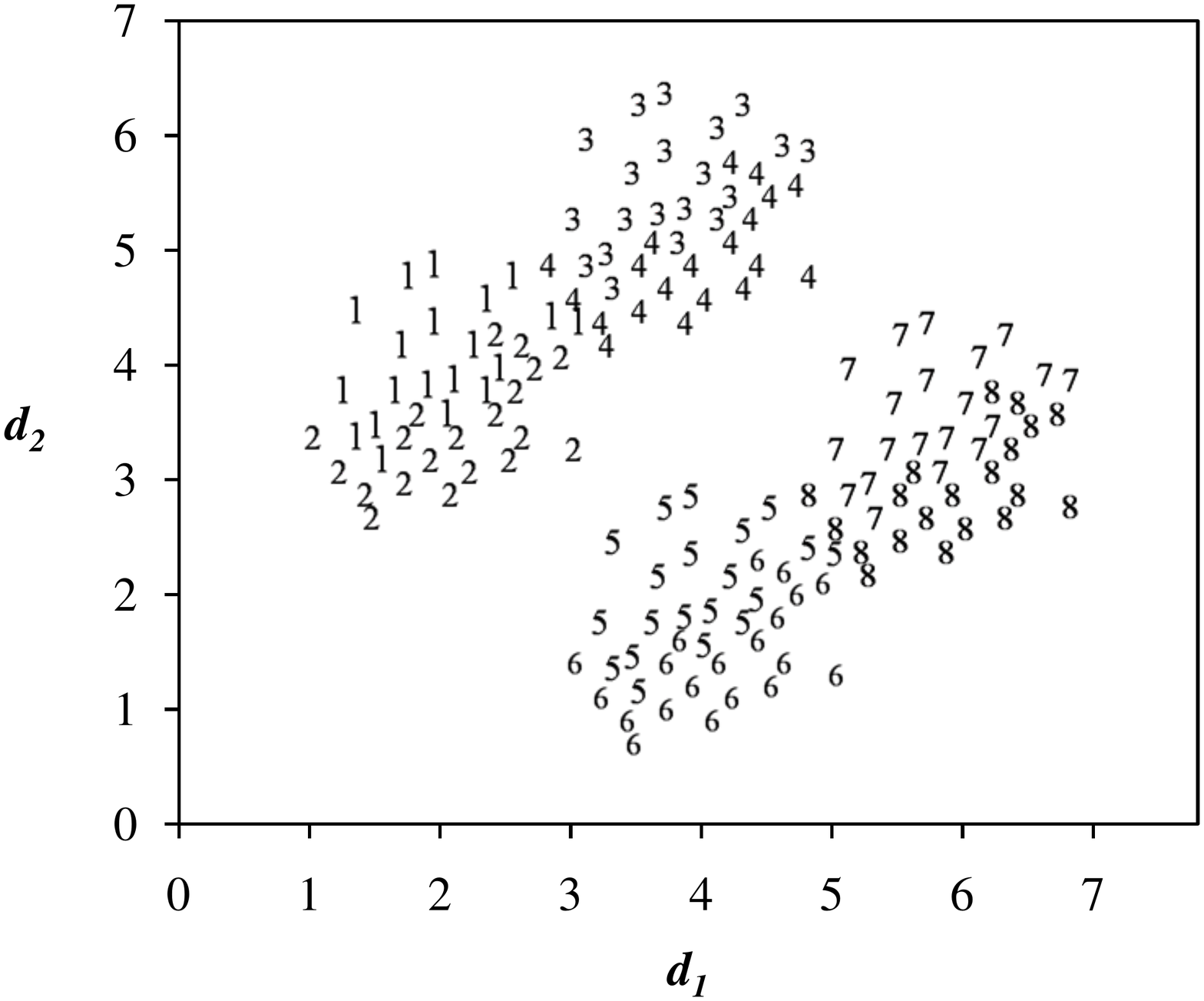}}
\caption{An example of two-dimensional artificial data including eight classes.}
\label{Arti_data}
\end{figure}

We aim to construct the code matrix in which each column is a suitable combination of positive and negative classes encoded to represent 
a binary model. 
As mentioned before, the best code matrix can be obtained by searching
from all possible $\frac{3^{N}-2^{N+1}+1}{2}$ columns for 
an $N$-class problem, and {thus} this is comparatively difficult when the number of classes increases.
{Our objective is to construct} the code matrix providing high quality of compression (the low number of binary classifiers) 
with high accuracy  of classification.
Although the highest compression of code can be possible by using $\lceil log_2N \rceil$ binary classifiers,
compression without proper combination of classes may lead to suffer with the complexity of classifier construction. 
{
To design a code matrix, for any classes i and j, some binary classifier(s) has to be  selected that constructs a pairing between a set of classes containing class i, and  another set containing class j.
}
We believe that the most important pairings affecting the classification accuracy are the 
pairs of classes with hard separation.
These pairings cannot be avoided as they must be included in some combinations of classes to distinguish between each other.
The number of classes combined in each classifier is varied from 2 to $N$ classes.
If we do not construct a classifier to separate between 
pair difficult classes,
to distinguish the two classes, we still have to build a classifier by using  other classes together with these two classes that increases the complexity of classifier construction.
For example, Fig.~\ref{Arti_data} shows that if we build the model with a linear function to distinguish 
classes 1 vs 2, which are hard-to-separate classes, by setting classes 1,3,4,7,8 as positive classes
and classes 2,5,6 as negative classes, it will not be easy to learn a good hyperplane.

\begin{algorithm*}[t]
  \caption{Optimal matching of subset of classes.}\label{CCSP}
  \begin{algorithmic}[1]
  \Procedure{Optimal matching of subset of classes.}{}
  \State Initialize set of classes $C \gets \{1, 2, 3, ..., N\}$, and set of subset of classes $S \gets \{S_1, S_2, S_3, ..., S_N\}$ 
         where $S_m$ $\gets$ $\{m\}$ and $m \in C$.
  \State Construct binary models of all possible pairs of $S_{p}$ and $S_{n}$ as a subset of classes with the positive label 
         and the negative label respectively, where  $1 \leq p < n \leq |S|$. 
  \State Calculate generalization performances of all possible pairs of $S_{p}$ and $S_{n}$ by using $k$-fold cross-validation
as shown in Algorithm~\ref{CV_algo}.
  \Do
        \State Apply the minimum weight perfect matching~\cite{Cook97} to find the optimal. 
		$\lfloor \frac{|S|}{2} \rfloor$ pairs of subset of classes from all possible 
		pairs on $S$ to obtain the closest pair of subset of classes based on the minimum generalization performance.
		\State Add all of optimal pairs as binary classifiers into the columns of the code matrix.
		\State $S \gets \emptyset$
		\State Combine each pair of subsets of classes into the same subset and then add it into $S$ as the new subset of classes.		
		\State Construct binary models of all possible pairs of $S_{p}$ and $S_{n}$. 
		\State Calculate generalization performances of all possible pairs of $S_{p}$ and $S_{n}$. 
  \doWhile{$|S|> 2$} 
  \State Add the last remaining two subsets as binary classifier  into the last column of the code matrix.
  \State \textbf{return} {\it code matrix}.
  \EndProcedure
  \end{algorithmic}
\end{algorithm*}

\begin{algorithm*}[t]
  \caption{An estimation of the generalization performance of a classifier by using $k$-fold cross-validation.}\label{CV_algo}
  \begin{algorithmic}[1]
  \Procedure{Cross Validation}{}
  \State Set of training data $T$ is partitioned into $k$ disjoint equal-sized subsets.
   \State Initialize the number of examples with correct classifications of round $i$: $w_i \gets 0$. 
  \For{$i$=1 to $k$ }
     \State {\it validate set} $\gets$ $i^{th}$ subset.  
     \State {\it training set} $\gets$ all remaining subsets.
     \State Learn model based on {\it training set}.
     \State $w_i$ $\gets$ Evaluate the learned model 
								by {\it validate set}, and find the number of examples with correct classifications.       
	 	 
  \EndFor  
  \State {\it generalization performance} $\gets$ $\sum_{i=1}^{k}w_i \times \frac{1}{|T|}$. 
  \State \textbf{return} {\it generalization performance}.
  \EndProcedure
  \end{algorithmic}
\end{algorithm*}

\begin{figure}
        \centering
        \begin{subfigure}[b]{0.22\textwidth}      
				\centering                          
                \includegraphics[width=\textwidth]{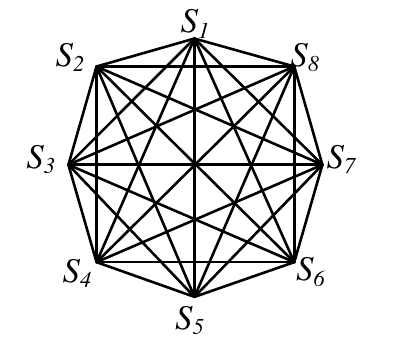}
                \caption{}
        \end{subfigure}%
        \begin{subfigure}[b]{0.22\textwidth}
				\centering
                \includegraphics[width=\textwidth]{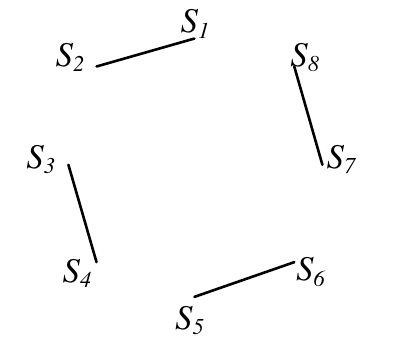}
                \caption{}
        \end{subfigure}             
        \caption{An example of the output of the minimum weight perfecting matching algorithm: 
		(a) all possible matchings of eight subsets of classes
		and (b) an output of the matching algorithm.}
		\label{Perfect_matching}
\end{figure}

\begin{figure}[htbp]
\centering
\scalebox{0.8}{\includegraphics{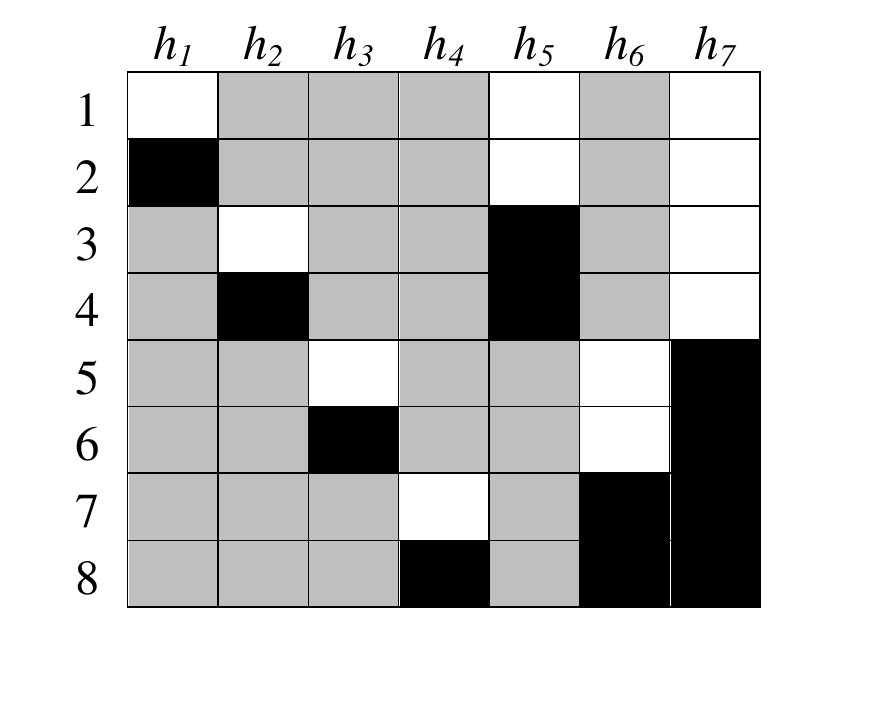}}
\vspace{-10pt}
\caption{An example of the code matrix constructed by the proposed method
based on Linear Support Vector Machines with $C$ = 1
for the artificial data from Fig.~\ref{Arti_data} 
(The white region, the dark region, and the gray region represent 
$\textquoteleft$1', $\textquoteleft$-1', and $\textquoteleft$0', respectively).}
\label{code_matrix}
\end{figure}

{
By the above reason, we carefully design the code matrix by considering the pairs of classes with hard separation 
as the first priority pairings 
and these pairings also indicate
the high similarity between the classes.
Intuitively, the pair of classes with difficult separation are allowed to combine with the low number of classes
in order to avoid situation mentioned before, and we expect that the easier pairing can be combined 
with the large number of classes without affecting the classification ability much.  
Based on our idea, the classes with high similarity are grouped together as the same class label, and the classes with low similarity are separated. 
For example, consider two-dimensional artificial data shown in Fig.~\ref{Arti_data}; classes 1,2,3,4 have high similarity and should be assigned with the same label (e.g. positive class),
and classes 5,6,7,8 also have high similarity and should be assigned with the same label (e.g. negative class).
Additionally, these two groups have low similarity by observation, 
and thus if we learn the binary model with a linear function, we will obtain absolutely separable hyperplane.
}

{
In order to obtain the code matrix satisfying the above requirements,
we employ the minimum weight perfect matching algorithm~\cite{Cook97} applied to find the optimal pairs of subset of classes
by using the generalization performance~\cite{Bartlett99} as a weighting criterion.
Our method is called {\it optimal matching of subset of classes} algorithm described in Algorithm~\ref{CCSP}.
}

For solving the optimal matching problem, let $G = (V, E)$ be a graph with node set $V$ and edge set $E$. Each node in $G$
denotes one subset of classes and each edge indicates one binary classifier of which 
generalization performance is estimated from Algorithm~\ref{CV_algo} (see Fig.~\ref{Perfect_matching}(a)).
The output of the matching algorithm for graph $G$ is a subset of edges with the
minimum sum of generalization performances of all edges and each node in $G$ is met by
exactly one edge in the subset (see Fig.~\ref{Perfect_matching}(b)). 

Given a real weight $w_e$ being generalization performance for each edge $e$ of $G$, the problem of matching
algorithm can be solved by the minimum weight perfect matching that finds a perfect
matching $M$ of minimum weight $\sum(w_e : e \in M)$.

%

For $U \subseteq V$, let $E(U) = \{(i,j):(i,j) \in E, i \in U, j \in U\}$.
$E(U)$ is the set of edges with both endpoints in $U$. The set of edges incident to
node $i$ in the node-edge incidence matrix is denoted by $\delta(i)$. 
The convex hull
of perfect matchings on a graph $G = (V, E)$ with $|V|$ even is given by
\\ \hspace*{2em}a) $x \in \{0,1\}^m$
\\ \hspace*{2em}b) $\sum_{e \in \delta(v)}x_e = 1$ for $v \in V$
\\ \hspace*{2em}c) $\sum_{e \in E(U)}x_e \leq \lfloor \frac{|U|}{2} \rfloor$ for all odd sets $U \subseteq V$ with $|U| \geq 3$
\\where $|E| = m$, and 
$x_e = 1$ ($x_e = 0$) means that $e$ is (is not) in the matching.
Hence, the minimum weight of a perfect matching (mp) is at least as large as the following value.

\begin{equation}
mp = min \sum_{e \in E} w_e x_e
\label{eqMWPM}
\end{equation}
\noindent where $x$ satisfies (a), (b), and (c). 
Therefore, the matching problem can be solved by the integer program in Eq.~(\ref{eqMWPM}).


{
On each round of Algorithm~\ref{CCSP}, we consider a sub-problem including $|S|$ subsets of classes, 
and calculate optimal pairs of subsets of classes by using the minimum weight perfect matching algorithm. 
The cost function employed in the minimum weight perfect matching algorithm is the sum of generalization performance
calculated from Algorithm~\ref{CV_algo}~\cite{Mitchell97}.
The obtained subsets of classes are then used as a column of the code matrix. 
Consider data on Fig.~\ref{Arti_data} 
and the designed code matrix in Fig.~\ref{code_matrix}, 
at the first round,
each element of $S$ contains only set of one class, and the size of $S$ is 8.
After applying the minimum weight perfect matching algorithm,
we get four optimal pairings, i.e., classifier 1 vs 2, classifier 3 vs 4,
classifier 5 vs 6, and classifier 7 vs 8.
All of optimal four pairs as binary classifiers are added into the columns of the code matrix, 
i.e., classifiers $h_1$ to $h_4$ 
(see Fig.~\ref{code_matrix}). 
The set $S$ is re-assigned by the empty set.
The classes from each pairing are combined together  
into the same subset and then are added into $S$ 
as new members.
Currently, the size of $S$ is 4, so we continue the next round. 
After applying the minimum weight perfect matching algorithm,
we get two optimal pairings, i.e., classifier 1,2 vs 3,4, and
classifier 5,6 vs 7,8.
These two optimal pairs as binary classifiers are added into the columns of the code matrix, 
i.e., classifiers $h_5$ to $h_6$.
The set $S$ is re-assigned by the empty set again, and then 
the classes from each pairing are combined together  
into the same subset and then are added into $S$ as new members.
Now, the size of $S$ is 2, so we exit the loop.
After that, the last remaining two subsets 
(the first subset including classes 1,2,3,4 and 
the second subset including classes 5,6,7,8) are added 
into the last column of the code matrix as binary classifier, i.e., classifier $h_7$.
}

In our methodology, a pair of classes that already are paired will not be considered again.
In the later round, the number of the members in $S$ increases.
For example at the last round, $S_1 \gets \{1,2,3,4\}$ and $S_2 \gets \{5,6,7,8\}$. 
It seems that this leads to complexity for calculating the hyperplane to separate between $S_1$ and $S_2$.
However, the remaining classes that can be paired trend to 
allow easier separation
due to the combination of classes in the same subset of classes 
having high similarity 
(in Fig.~\ref{Arti_data}, classes 1,2,3,4 are very close, and classes 5,6,7,8 are also close),
while the possible sixteen pairings including 1 vs 5, 1 vs 6, 1 vs 7, 1 vs 8, 2 vs 5, ..., 4 vs 8 contain classes with high dissimilarity.
It illustrates that these sixteen pairings are encoded in only one binary classifier 
as the One-Versus-One requires sixteen binary classifiers.
This example also confirms that based on our technique although the number of classes for constructing a binary model increases, 
each subset of positive and negative classes is combined appropriately, and this leads to effective separation.

\section{Experiments}\label{experiments}  

In this section, we design the experiment to evaluate the performance 
of the proposed method. We compare our method with the classic codes, 
i.e., the dense random code, the sparse random code~\cite{Dietterich95,Allwein2000}, and the One-Versus-One~\cite{Hastie98}.
This section is divided into two parts as experimental settings, and results \& discussions.

\vspace*{-0.5em}\subsection{Experimental Settings}\label{ExpProtocol}
We run experiments on ten datasets from the UCI Machine Learning
Repository~\cite{Blake98}.
For the datasets containing both training data and test data, 
we added up both of them into one set, and used 5-fold cross validation 
for evaluating the classification accuracy.

\begin{table*}[htp]
\centering
\caption[]{{Description of the datasets used in the experiments.}}
\begin{tabular}{l c c c}
\hline
\textbf{Datasets} & \textbf{\#Classes} & \textbf{\#Features}& \textbf{\#Cases}\\
\hline
Segment & 7 & 18& 2,310  \\
Arrhyth	&	9&	255&438\\
Mfeat-factor&10&216&2,000\\
Optdigit&10&62&5,620\\
Vowel&11&10&990\\
Primary Tumor&13&15&315\\
Libras Movement &15&90&360\\
Soybean&15&35&630\\
Abalone&16&8&4,098\\
Spectrometer&21&101&475\\
\cline{1-4}
\end{tabular}
\label{Dataset_Dessciption}
\end{table*}

\begin{table*}[htp]
\centering
\caption[]{{ A comparison of the classification accuracies by using the dense random code and the proposed method. }}
{
\begin{tabular}{l l l l }
\hline
{Data sets} & {Dense random code}& \hspace*{1em} &{Proposed method}\\
\hline
Segment         &91.018$\pm$0.112           									&{ }&\textbf{92.857$\pm$0.159*}\\
Arrhyth         &\textbf{64.626$\pm$0.246*}                       				&{ }&60.731$\pm$0.298\\
Mfeat-factor    &\textbf{97.073$\pm$0.228}           							&{ }&97.050$\pm$0.188\\
Optdigit        &99.034$\pm$0.056                              					&{ }&\textbf{99.093$\pm$0.075}\\
Vowel   		&59.632$\pm$0.980 					          					&{ }&\textbf{80.808$\pm$0.479*}\\
Primary Tumor	&43.931$\pm$0.567 					          					&{ }&\textbf{44.127$\pm$1.085}\\
Libras Movement 		&82.507$\pm$0.372                               				&{ }&\textbf{84.722$\pm$0.621*}\\
Soybean         &93.273$\pm$0.418                               				&{ }&\textbf{93.651$\pm$0.251}\\
Abalone         &23.126$\pm$0.012                               				&{ }&\textbf{26.940$\pm$0.152*}\\
Spectrometer    &50.428$\pm$0.420           									&{ }&\textbf{57.474$\pm$1.080*}\\
\hline
\end{tabular}}
\label{Dense_ProposedM_comparison}
\end{table*}

\begin{table*}[htp]
\centering
\caption[]{{ A comparison of the classification accuracies by using the sparse random code
and the proposed method. }}
{
\begin{tabular}{l l l l }
\hline
{Data sets} & {Sparse random code}& \hspace*{1em} &{Proposed method}\\
\hline
Segment         	&91.937$\pm$0.105           									&{ }&\textbf{92.857$\pm$0.159*}\\
Arrhyth         	&\textbf{62.112$\pm$0.225*}                       				&{ }&60.731$\pm$0.298\\
Mfeat-factor    	&97.048$\pm$0.212		           								&{ }&\textbf{97.050$\pm$0.188}\\
Optdigit        	&99.072$\pm$0.059                              					&{ }&\textbf{99.093$\pm$0.075}\\
Vowel   			&61.986$\pm$0.951 					          					&{ }&\textbf{80.808$\pm$0.479*}\\
Primary Tumor		&\textbf{44.770$\pm$0.864} 					          			&{ }&44.127$\pm$1.085\\
Libras Movement 	&83.421$\pm$0.311                               				&{ }&\textbf{84.722$\pm$0.621}\\
Soybean         	&93.115$\pm$0.421                               				&{ }&\textbf{93.651$\pm$0.251}\\
Abalone         	&24.311$\pm$0.023                               				&{ }&\textbf{26.940$\pm$0.152*}\\
Spectrometer    	&52.670$\pm$0.433           									&{ }&\textbf{57.474$\pm$1.080*}\\
\hline
\end{tabular}}
\label{Sparse_ProposedM_comparison}
\end{table*}

\begin{table*}[htp]
\centering
\caption[]{{A comparison of the classification accuracies by using the One-Versus-One and the proposed method. }}
{
\begin{tabular}{l l l l }
\hline
{Data sets} & {One-Versus-One}& \hspace*{1em} &{Proposed method}\\
\hline
Segment         &92.741$\pm$0.174           									&{ }&\textbf{92.857$\pm$0.159}\\
Arrhyth         &60.502$\pm$0.161                               				&{ }&\textbf{60.731$\pm$0.298}\\
Mfeat-factor    &\textbf{97.183$\pm$0.189}           							&{ }&97.050$\pm$0.188\\
Optdigit        &\textbf{99.170$\pm$0.062}                              		&{ }&99.093$\pm$0.075\\
Primary Tumor	&\textbf{44.550$\pm$0.831} 					          			&{ }&44.127$\pm$1.085\\
Vowel   		&\textbf{80.909$\pm$0.258} 					          			&{ }&80.808$\pm$0.479\\
Libras Movement 		&84.352$\pm$0.317                               				&{ }&\textbf{84.722$\pm$0.621}\\
Soybean         &92.989$\pm$0.422                               				&{ }&\textbf{93.651$\pm$0.251}\\
Abalone         &26.326$\pm$0.063                               				&{ }&\textbf{26.940$\pm$0.152}\\
Spectrometer    &57.193$\pm$0.775           									&{ }&\textbf{57.474$\pm$1.080}\\
\hline
\end{tabular}}
\label{OVO_ProposedM_comparison}
\end{table*}

\begin{table*}[htp]
\centering
\caption[]{{A comparison of the number of binary classifiers among the difference methods. }}
{
\begin{tabular}{l c c c c c }
\hline
{{Data sets}}&{\#classes} & {Dense}& {Sparse}& {One-Versus-One}  & {Proposed }\\
{{}}&{} & {random code }& {random code}& {}  & {method}\\
\hline
Segment         &7	&29	&43 &21 &6\\
Arrhyth         &9	&32	&48 &36 &8\\
Mfeat-factor    &10	&34	&50 &45 &9\\
Optdigit        &10	&34	&50 &45 &9\\
Vowel        	&11	&35	&52 &55 &10\\
Primary Tumor 	&13	&38	&56 &78 &12\\
Libras Movement		&15	&40	&59 &105 &14\\
Soybean         &15	&40	&59 &105 &14\\
Abalone         &16	&40	&60 &120 &15\\
Spectrometer    &21	&44	&66 &210 &20\\
\hline
\end{tabular}}
\label{time_comparison}
\end{table*}

In these experiments, we scaled data to be in [-1,1]. 
In the training phase, we used software package $SVM^{light}$ version 6.02~\cite{Joachims98,Joachims99}
to create the binary classifiers.
The regularization parameter $C$ of 1 was applied for model construction; 
this parameter is used to trade off between error of the SVM on training data and margin maximization.
We employed the RBF kernel $K(x_i,x_j) \equiv e^{-\gamma ||{\bf x_i}-{\bf x_j}||^2}$,
and applied the degree $\gamma=0.1$ to all datasets.  
{
}
For the dense random code and the sparse random code, ten code matrices were randomly generated
by using a pseudo-random tool~\cite{Bagheri2012}.
In case of sparse random codes, the probability 1/2 and 1/4 were applied to generate the bit of $\textquoteleft$0', 
and the other bits, i.e.. $\textquoteleft$1' and $\textquoteleft$-1', respectively.
For decoding process, we employed the attenuated euclidean decoding~\cite{Escalera2010} by
using $AED(x,y_i)=\sqrt{ \sum_{j=1}^{L}|y_{i}^{j}||x^j|(x^j-y_{i}^{j})^2}$ where $x$ is binary output vector 
and $y_i$ is a code word belonging to class $i$.

\vspace*{-0.5em}\subsection{Results \& Discussions}\label{ExpResult}
We compared the proposed method with three traditional techniques, i.e.,
the dense random code, the sparse random code, and  the One-Versus-One. 
{The comparison results are shown in Table~\ref{Dense_ProposedM_comparison} to Table~\ref{OVO_ProposedM_comparison}
in which all of datasets are sorted in ascending order by their number of classes.
}
For each dataset, the best accuracy among these algorithms is illustrated in bold-face and 
the symbol \textquoteleft*' means that the method gives the higher accuracy 
at 95 \% confidence interval with one-tailed paired t-test.

The experimental results in Table~\ref{Dense_ProposedM_comparison} shows that the proposed method yields
highest accuracy in almost all datasets.
The results also show that, at $95$\% confidence interval, the proposed technique performs
statistically better than the dense random code in five datasets and there is only one data set, i.e.,
the Arrhyth that the proposed method 
{ does not archive the better result.
}
The experimental results in Table~\ref{Sparse_ProposedM_comparison}
illustrates that the proposed method gives higher accuracy compared to the sparse random code in almost all datasets.
The results also show that, at $95$\% confidence interval, the proposed technique performs
statistically better than the sparse random code in four datasets and 
there is only one data set, i.e., the Arrhyth that the proposed method 
{does not provide the higher accuracy.} The combination method with a few binary classifiers may be weaker 
compared with one with the large number of binary classifier. 
Our proposed method is carefully designed to find the proper combination of classes,
and to create the binary classifiers with high efficiency.
However, as our method takes a few binary classifiers, it is possible that this situation may occur as found in case of the Arrhyth dataset.
Moreover, the classification performance also relates to the structure of the code matrix. 
We consider the size of search space that is proportional to $\frac{3^{N}-2^{N+1}+1}{2}$ 
where $N$ is the number of classes. The dense random code and the sparse random code which is independent to the selection of a code matrix, while our method tries to find a specific structure of code matrix according to 
the relation of generalization performance of their classes.
In case of the low number of $N$ with the small size of search space, the random techniques may obtain
a good solution while in case of the higher number of $N$ with the bigger size of search space
their probabilities to reach an expected solution decrease proportional to the growth of $N$.  
In this aspect, our method based on utilization of generalization performance as a guideline 
to find a solution is not affected.
{However, by the above reason in case of the small number of classes 
the code matrix generated by a random technique with the larger number of binary models 
may lead to the better classification performance as in the Arrhyth dataset mentioned before.}
Next, the last classification results in Table~\ref{OVO_ProposedM_comparison}
shows that the proposed method gives a little better results compared to the One-Versus-One in several datasets, and there is no significant difference between these two methods at $95$\% confidence interval.

{
Consider the characteristic of the obtained binary models by using our technique. Some binary classifiers include the large number of classes
that seem to be complex to construct binary models with high accuracy (as the first important issue).
For example, in case of the Spectrometer dataset including 21 classes,
twenty binary classifiers are obtained.
The number of classes containing in the binary classifier is varied from 2 to 21 classes.
For some hard-separation parings, our method allows the binary classifiers with the low number of classes to be combined
to avoid the situation mentioned in Section~\ref{Proposed_Method} (as the second important issue). 
The effects of these two issues to the designed code matrix can be observed via the classification accuracies.
Among all algorithms based on this dataset, our code matrix gives the higher classification accuracies 
compared to the dense random code and the sparse random code,
while providing a little better results compared to the One-Versus-One.
Moreover, the experimental results confirm that generally our algorithm provides the better code matrix
that the proper subsets of classes are combined.
}

{
We also compare the classification time (the number of binary classifiers employed) of the proposed method to the traditional works
as shown in Table~\ref{time_comparison}.
The results illustrate that the proposed technique requires comparatively low running time in all datasets
especially, when the number of classes is relatively large.
Our technique reduces large number of classification times compared to all previous works; the dense random code,
the sparse random code, and the One-Versus-One require $\lceil 10log_2N \rceil$, $\lceil 15log_2N \rceil$, 
and $N(N-1)/2$, respectively while our framework needs only $N-1$ for $N$-class problems.
}
{
Our code matrix construction requires the calculation 
of class matching using the minimum weight perfect matching that 
the estimation of generalization performances are employed as a weighting criterion.
Although this process needs additional computation to estimate the weight values using $k$-fold cross validation,  
the task is conducted in the training phase, and not affecting the performance in the classification phase.
}

\section{Conclusion}\label{conclusion}
{
We propose an algorithm to find the suitable combination of classes for creating the binary models 
in the ECOC framework by using the generalization performance as a relation measure among subset of classes. 
This measure is applied to obtain the set of the closest pairs of subset of classes 
via the minimum weight perfect matching algorithm in order to generate the columns of the code matrix. 
The proposed method gives higher performance both in terms of accuracy and classification times
compared to the traditional methods, i.e., the dense random code and the sparse random code.
Moreover, our approach requires significantly smaller number of binary classifiers 
while maintaining accuracy compared to the One-versus-One.
However, the expected matchings of subset of classes are not possibly available in some cases 
because relation of their subset of classes may force to combine inappropriate subset 
of classes and it may lead to misclassification.
We will further analyze to address this situation in our future work.
}

\section{Acknowledgment}\label{Acknowledgment}
This research is supported by the Thailand Research Fund, Thailand.


\end{document}